\documentclass[11pt]{article}
\usepackage[margin=1in]{geometry}
\usepackage{graphicx}
\usepackage[normalem]{ulem}
\usepackage{tikz-cd}

\usepackage{adjustbox}

\usepackage{amsmath,amssymb}

\usepackage{hyperref}
\hypersetup{
colorlinks = true,
linkcolor = violet,
citecolor = violet,
urlcolor = violet
}

\title{A Topological Approach for Motion Track Discrimination}

\author{
  Tegan Emerson$^{1}$, Sarah Tymochko$^{1,2}$, George Stantchev$^{3}$, Jason A. Edelberg$^{3}$,\\ Michael Wilson$^{3}$, and Colin C. Olson$^{3}$
}

\date{
$^1$Pacific Northwest National Laboratory, Seattle, WA, USA \\
$^2$Dept. of Computational Mathematics, Science and Engineering, Michigan State University, East Lansing, MI, USA \\
$^3$US Naval Research Laboratory, Washington, DC, USA \\
}

\begin{document}

\maketitle

\begin{abstract}
 Detecting small targets at range is difficult because there is not enough spatial information present in an image sub-region containing the target to use correlation-based methods to differentiate it from dynamic confusers present in the scene. Moreover, this lack of spatial information also disqualifies the use of most state-of-the-art deep learning image-based classifiers. Here, we use characteristics of target tracks extracted from video sequences as data from which to derive distinguishing topological features that help robustly differentiate targets of interest from confusers. In particular, we calculate persistent homology from time-delayed embeddings of dynamic statistics calculated from motion tracks extracted from a wide field-of-view video stream. In short, we use topological methods to extract features related to target motion dynamics that are useful for classification and disambiguation and show that small targets can be detected at range with high probability.
\end{abstract}

\section{Introduction}

The ability to identify and label objects in imagery and video has long been of interest to communities across the civilian and military sectors. As higher resolution imaging systems have evolved, there has been corresponding growth in the development and use of automated algorithms to perform these identification tasks. Recently, deep learning methods have demonstrated dramatic improvements in both identification and classification tasks for data arising from applications ranging from self-driving cars, natural language processing, and image classification \cite{bojarski2017explaining,radford2019language,iandola2016squeezenet}. Due to their success, deep learning algorithms have become the default technique when presented with high-volume, high-resolution image analysis tasks. However, not all imaging challenges are appropriate for Deep Learning techniques; in particular, in cases where labelling data is difficult or where high-resolution sensors (e.g., in the infrared domain) are expensive or unavailable.

We consider here the problem of detecting small targets at range. This is a difficult task to automate because of two competing requirements: (1) wide field-of-view (WFOV) cameras are required for sufficient coverage of a scene but (2) high resolutions are needed in order to accurately detect and classify objects moving within the scene. The latter requirement is especially critical because the scene may contain moving objects whose tracks appear similar to the target's at the given resolution, thus registering as false alarms if classification decisions are made solely based on change detection. Such objects are referred to as dynamic confusers.
In addition, The WFOV requirement, creates a scenario where there is generally not enough spatial information present in an image sub-region containing a possible detected target to use correlation-based methods to differentiate it from confusers.

As a result, rather than rely on correlation-based methods (e.g., classification via convolutional neural networks) we instead use characteristics of target tracks extracted from video sequences as data from which to derive distinguishing topological features that clearly identify target class. In essence, we use topological methods to extract features related to target movement dynamics that are useful for classification and disambiguation.

We present here a novel approach for differentiating between targets and confusers based on an analysis of motion tracks derived from object and change detection algorithms. Our approach performs well as a stand-alone technique on the test set considered but could also be considered as a preprocessing step prior to a deep learning implementation. The approach leverages mathematical techniques for time series analysis (i.e., time-delayed embeddings) to ensure that informative dynamic invariants are preserved such that a topological analysis of the resulting point cloud data yields features that are useful for classification.

Section \ref{sec:background} provides an introduction to the existing tools and techniques we utilize (i.e., time-delay embedding, topological data analysis, and motion track generation). Next, in Section \ref{sec:data} we provide a description of the data set, discuss the experimental design and how we address the pre-processing data challenges. In Section \ref{sec:approach} we outline our framework for analyzing motion tracks. Experimental results are presented in Section \ref{sec:experiment}, followed by discussion of the results and the implications for future work in Section \ref{sec:conclusion}.

\section{Background}
\label{sec:background}
\subsection{Time-Delay Embedding}
\begin{figure}[htbp]
\centering
\includegraphics[width=.7\textwidth]{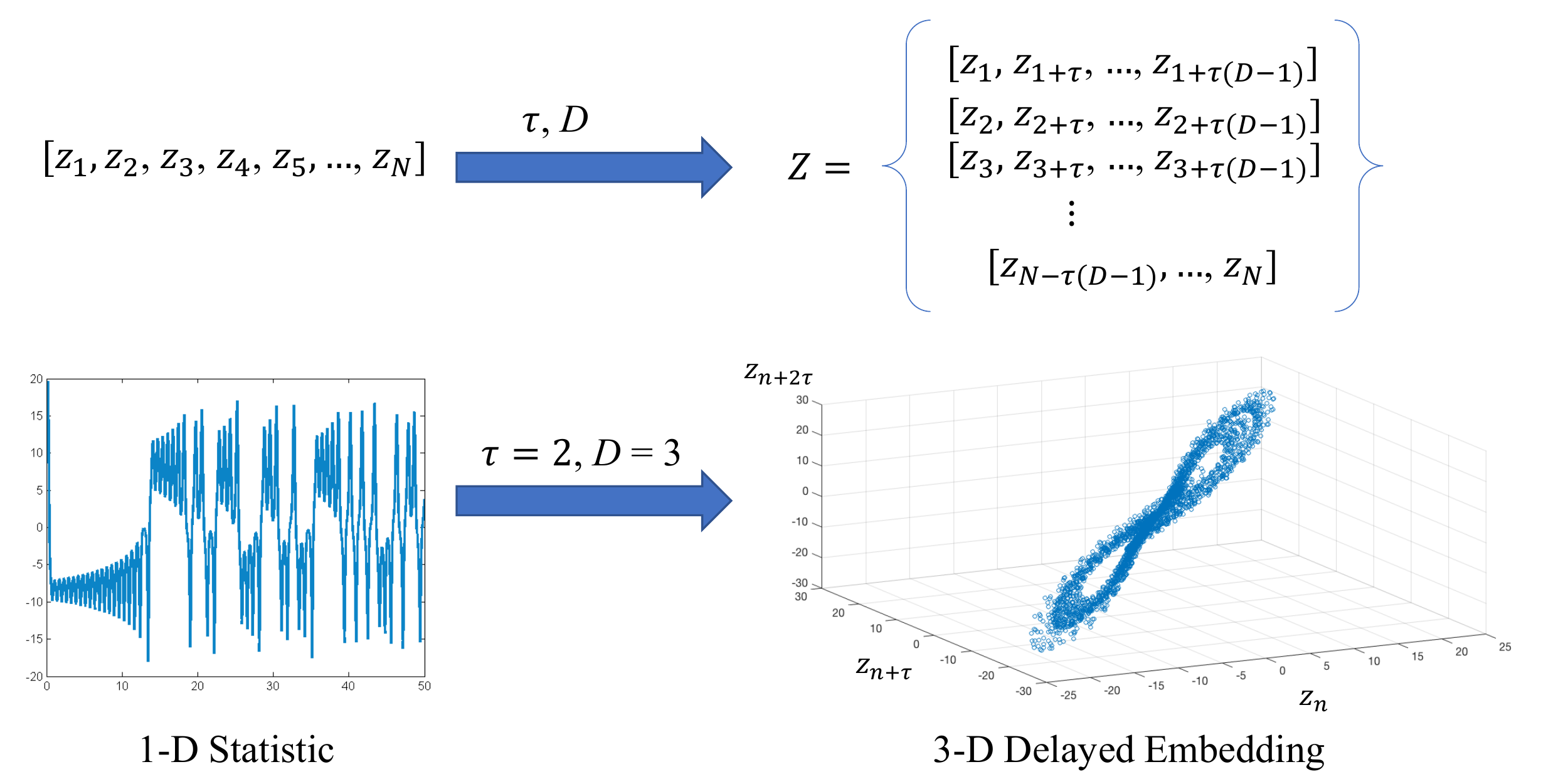}
\caption{Example of how a point cloud is created from a $1$-dimensional statistic. We define the notation generally and visually demonstrate the result for a given $1$-dimensional statistic, a delay of $\tau=2$, and an embedding dimension of $D=3$. }
\label{fig:TDE}
\end{figure}

Time-Delay Embedding (TDE) is a technique for time-domain signal analysis based on the seminal work of Takens \cite{Takens:81} and Whitney \cite{Whitney:36}. The underlying assumption of TDE is that, given a dynamical system describing a physical process, some measured aspect of the system is a time-varying statistic from which certain uniquely identifying dynamic invariants may be extracted given a ``proper'' embedding of the observed statistic. Such an embedding is constructed in turn by using algorithms such as false nearest neighbors \cite{Kennel:92} and mutual information \cite{Fraser:86} to select, respectively, an appropriate embedding dimension, $D$, and time-delay, $\tau$, to apply to an observed time series, $\vec{z}=\{z_{n}\}_{n=1}^{N}$, such that $Z=\{(z_{n}, z_{n+\tau}, \cdots, z_{n+(D-1)\tau}) \}_{n=1}^{N-(D-1)\tau}$ . Thus, the output of a TDE is a cloud of $D$-dimensional points constructed from delayed copies of the observed $1$-dimensional time series as shown in Fig. \ref{fig:TDE}.

In the context of the target vs. confuser problem, this translates to the assumption that the motion patterns of each object can, in theory, be modeled by a dynamical system and that the motion of an object as projected onto the sensor is the observed dynamic statistic. Even though some dynamic information has been lost due to this projection, Takens's theorem guarantees that certain dynamic invariants describing the topology of the underyling manifold on which the dynamics evolve will still be preserved and observable via the delay embedding process. Given this preservation of topological invariants, we leverage tools from topological data analysis to help generate uniquely identifying features from the projected motion tracks.

\subsection{Topological Data Analysis}
Topological Data Analysis (TDA) describes a broad collection of techniques which aim to extract invariants of and/or characterize data based on the mathematical notion of ``shape'' or ``connectedness''. The use of TDA has grown in recent years due to the flexibility of the toolset and increased computational e{ff}iciency. In this section we provide a brief high-level description of two TDA tools: Persistent Homology (PH) and Persistence Images (PIs). We omit the use of technical formulations and refer the interested reader to \cite{Carlsson:2008} and \cite{Edelsbrunner:2008} for formal definitions and to \cite{Otter:2017} for algorithms and implementation.

\subsubsection{Persistent Homology}
 Homology in general can be intuitively understood as an algebraic framework for consistently counting distinct classes of ``holes'' and connected components of various dimensions present in a topological space. Technically, a $k$-dimensional hole is a subset of a $D$-dimensional space that can be deformed continuously (i.e., without tearing) to a $k$-dimensional sphere $S^k$ inside the space but cannot be continuously deformed to a point. In other words, a $k$-dimensional hole is \emph{topologically equivalent} to $S^k$ inside the space. For example, a 0-dimensional hole is trivially any subset that can be continuously deformed to a point. A 1-dimensional hole is topologically equivalent to a simple loop (a circle) and a 2-dimensional hole is topologically equivalent to the usual 2-dimensional sphere.

 Visualizing holes for higher dimensions becomes less intuitive but the notion is mathematically well-defined and generalizes directly. Two $k$-dimensional holes are equivalent (belong to the same homology class) if they be continuously deformed to each other inside the space. For instance, the two-dimensional torus (the boundary of a doughnut) has two classes of loops -- one that is equivalent to a longitudinal circle and another one which is equivalent to a latitudinal circle.

 Homology can be used to analyze data, that is, a finite collection of points in a $D$-dimensional space, by associating it with a specific combinatorial topolological space whose homological structure (the algebraic relationship among the equivalence classes of holes at various dimension) can be computed \cite{Edelsbrunner:2010}. In particular, the number of equivalence classes of $k$-dimensional holes at a given scale, is referred to as the $k^{th}$ Betti number \cite{Ghrist:2014}. This association is scale-dependent and consequently so would be the resulting homological structure. Persistent Homology (PH) is a common tool in TDA that aims at extracting and encoding scale-dependent topological information from data by tracking how the homology structure of certain topological spaces associated with the data varies as a function of the scale parameter \cite{Edelsbrunner:2008}.

\begin{figure}[htbp]
\centering
\includegraphics[width=0.9\textwidth]{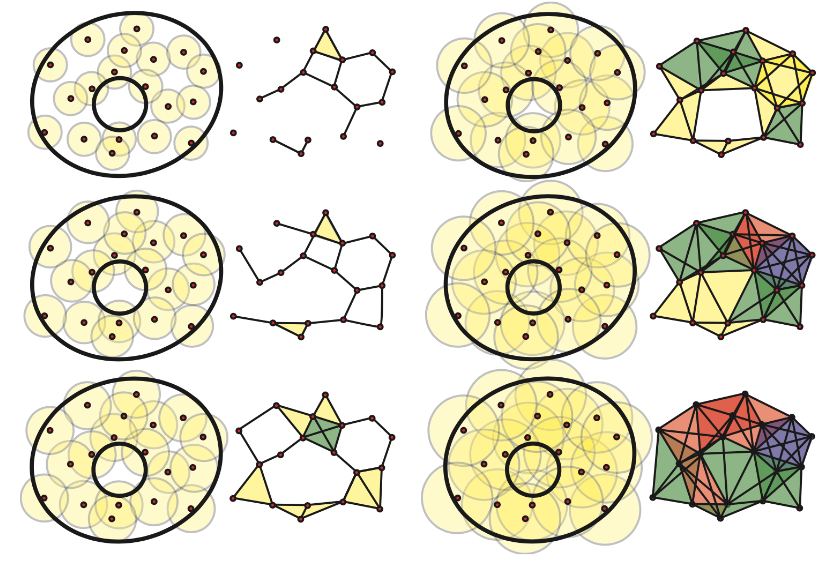}
\caption{A sequence of Vietoris Rips complexes for a point cloud data set representing
an annulus. Upon increasing the radius of the balls, we see holes appear and disappear.
Image reproduced from \cite{Ghrist:2008}}
\label{fig:VRcomplex_Ghrist}
\end{figure}

\begin{figure}[htbp]
\centering
\includegraphics[width=0.9\textwidth]{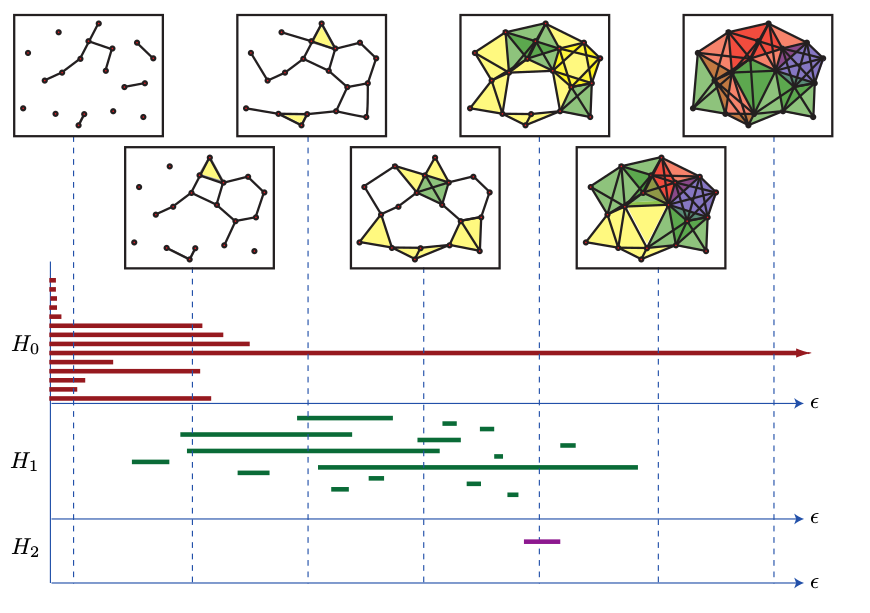}
\caption{An example of a barcode diagram for dimensions 0, 1, and 2 of the filtered simplicial complex in Fig.~\ref{fig:VRcomplex_Ghrist}. The $\varepsilon$ parameter corresponds to the ball radius and controls the structure of the associated Rips Complex. Image reproduced from \cite{Ghrist:2008}}
\label{fig:VRcomplex+BCD}
\end{figure}

PH can be computed in a variety of ways depending on the data under consideration. For our approach we will be computing the PH of a point cloud in a real-valued Euclidean space. Within this context, scale is equivalent to distance, which for a fixed value can be used to construct a topological space called the Vietoris-Rips (VR) complex \cite{Edelsbrunner:2010}. The VR complex is formed by adding edges, faces, tetrahedral cells, and so forth based on the pairwise intersections of balls of radius equal to the chosen distance parameter as shown in Fig.~\ref{fig:VRcomplex_Ghrist}. The Betti numbers of the associated VR complex can then be computed using linear algebra. Varying the distance parameter and evaluating the associated homology structure allows the tracking of scales across which each equivalence class of $k$-dimensional holes \emph{persists}. The collection of intervals, the end points of which represent respectively the birth/disappearance of a homological feature, is called a \emph{Barcode Diagram} (see Fig.~\ref{fig:VRcomplex+BCD}).
For each dimensional homology we can summarize the information in a variety of ways. In early TDA approaches each homology dimension was typically summarized in a Persistence Diagram (PD)--a multiset of points in the plane where the $x-$value of a point corresponds to the scale at which the hole appeared and the $y-$value indicates the scale at which the hole disappeared/collapsed (see for instance \cite{Edelsbrunner:2010}). The space of PDs can be endowed with a metric structure allowing one to measure the similarity between point clouds. However, the metrics on PDs are typically expensive to compute and tend to limit the tools available to differentiate between configurations of point clouds. Therefore, significant effort has been spent in recent years on developing alternative representations of the information captured in a PD such that broader machine learning techniques can be applied. One such representation, a Persistence Image (PI), is described next.

\subsubsection{Persistence Images}
Persistence images are generated by centering 2-dimensional probability distributions on each point in a PD to form a surface, overlaying a grid, and then integrating the area under the surface over each discretized region of the grid. See \cite{PIs} for the steps involved in forming a PI. Typically the 2-dimensional distribution is chosen to be Gaussian requiring a choice of covariances. The other associated parameter is the size of the grid. It has been shown \cite{PIs,3DSurface} that the performance of PIs is robust to the choice of these parameters. For our implementation we chose a grid size/resolution of 25 and the default parameter settings for the PI code found at \cite{ripser}.

\section{Track Generation and Conditioning}
\label{sec:data}
The data we consider are acquired using a stationary wide field-of-view system which captures images of the same scene over time. Images are collected using a commercial camera. Using custom change detection algorithms we are able to track objects as they move within or across the field-of-view as shown in Fig. \ref{fig:dataSample}.

The resolution of the imaging system is sufficient to detect movement but, as can be seen from the zoomed-in image clips in Fig. \ref{fig:dataSample}, is not high enough to resolve the moving objects. It is possible to do targeted high-resolution imaging on objects of interest, but without further processing the number of nontrivial confusers is prohibitive. The experiment and results we present in the subsequent section have been considered as standalone, however, it is plausible to employ the presented methodology as a first step to reduce the number of false alarms to such an extent that targeted imaging could then be employed to extract images on which a deep learning framework could be applied.

\subsection{Motion Track Generation}

Track data are derived from visible, monochromatic, imagery that contains dynamic targets of interest as well as dynamic confusers. Imagery is processed frame-by-frame using a custom target detection processor. In particular, raw detections are generated via a combination of temporal and spatial processing, the results of which are fused to allow the continued detection of slow-moving targets while simultaneously providing a live adaptive non-uniformity correction of the imaging system.

These frame-by-frame detections are then correlated in space and time resulting in individual track files corresponding to each moving object in the image series. The target tracks are then determined via comparison with ground truth data. Finally, an operator reviews the source imagery and track information to separate the remaining non-target tracks into different confuser classes.

\subsection{Motion Sub-Track Generation}
\label{sec:subtrack}
\begin{figure}[htbp]
\centering
\includegraphics[width=.8\textwidth]{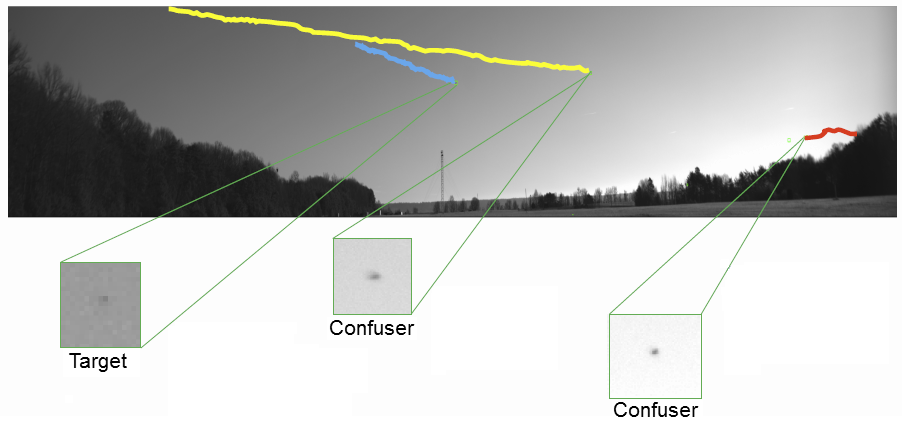}
\caption{Example of analyzed data. The image chips contained in each green box correspond to a zoomed-in view, at a single frame, of the object generating the corresponding motion track.}
\label{fig:dataSample}
\end{figure}
\begin{figure}[htbp]
\centering
\includegraphics[width=.65\textwidth]{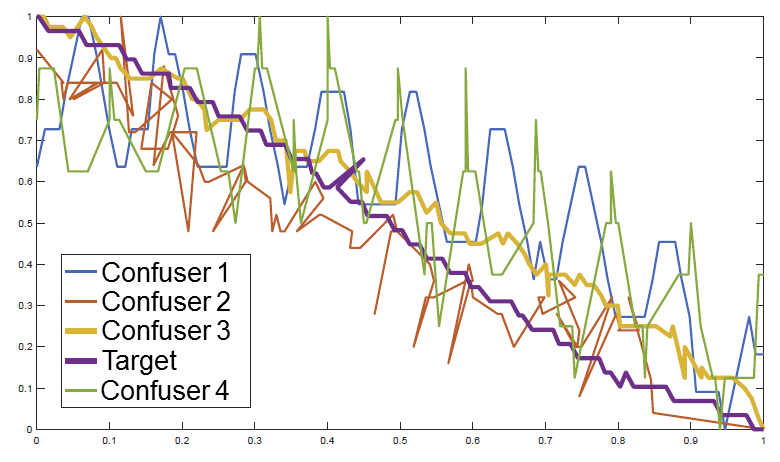}
\caption{Examples of normalized sub-tracks taken from each of the five test objects.}
\label{fig:subtracks}
\end{figure}

Fig. \ref{fig:dataSample} further highlights another challenge for analysis of motion tracks: tracks of differing lengths. Not all objects appear in the image at the same time and nor do they stay in the field-of-view for the same number of frames. To address this issue we generate sub-tracks of the same length for all tracks which allows for an appropriate comparison between all tracks. Here, the analyzed data set consists of four confuser motion tracks and one target track. The five evaluated tracks each have a ground-truth label.

For each track, $t_{j}$, we have a set of $(x,y)$-coordinate pairs corresponding to the location in the image indexed by time, that is,  $t_{j}=\{(x^{j}_{i},y^{j}_{i})\}_{i=1}^{N_{j}}$ where $i$ counts the number of frames and $N_{j}$ is the total number of frames for which the $j^{th}$ object is detected. Let $N^{*}<N_{j}$ be a desired sub-track length. For each track we generate the $k^{th}$ sub-track of the $j^{th}$ object  as $s_k^{j}=\{(x^{j}_{(k-1)+i},y^{j}_{(k-1)+i})\}_{i=1}^{N^{*}}$. Each sub-track is a subset of its parent track and all sub-tracks are the same length $N^{*}$ such that they may be directly compared both within and across classes. In short, the $k^{th}$ sub-track is an $N^{*}$-length window extracted from the parent track and the $(k+1)^{th}$ sub-track is generated by shifting the extraction window by one. Larger time shifts are possible but are not investigated here.

Because an object needs to retain its dynamic representation regardless of the initial location within the field of view, we normalize each sub-track such that $\tilde{s}_{k}^{j}\in [0,1]\times[0,1]$. We perform this normalization by first subtracting the minimum value from within a sub-track from all elements composing the sub-track and then divide that result by the maximum value of the result to force all elements to map to the range $[0,1]$. This normalization is performed independently for both the $x$- and $y$-components comprising the sub-track.  Fig. \ref{fig:subtracks} shows a single, normalized sub-track from each of the five original tracks. Notice that while three of the confuser sub-tracks are visually distinct from the target, the third confuser is more similar to the target than to the other confusers.

Finally, we generate a vector of statistics from a sub-track by projecting 2-dimensional track coordinates into a 1-dimensional vector representation by calculating $\vec{s}_k^{j}=\tilde{s}_k^{j}\cdot v$ where $v$ is a $2$x$1$ vector with random entries drawn uniformly from $(0,1]$. That random vector is formed once, remains fixed, and is applied to all sub-tracks from all objects in the scene. All told, given $N_{j}$ and a choice of $N^{*}$, we produce a set, $S_{j}$, comprised of $K_{j}=(N_{j}-N^{*})$ vectors $\vec{s}_k^{j}$ for each moving object in the scene.  The generation of normalized sub-tracks and their associated statistics completes the preprocessing stage and yields the data on which we perform our experiments.

\section{Topological Approach for Motion Tracks}
\label{sec:approach}

In many situations the exact variables that are needed to build explicit, accurate dynamic models for different systems or processes are unknown or the number required is prohibitive. For some applications the technology needed to capture relevant measurements may not even exist. As a result there is a need for other means of analyzing signals/data that \emph{can} be acquired. Takens's Theorem, discussed in Section \ref{sec:background}, provides a rigorous mathematical framework for doing exactly this, which we can leverage for disambiguating target dynamics from those of confusers.

To do so, we assume that the dynamic systems governing the motion mechanics of confusers differ from those governing the targets of interest. Moreover, we assume that the motion pattern/shape projected through the field of view of the sensor is a representative statistic of system dynamics. Finally, we assume that the dynamical trajectories of the underlying systems evolve on different manifolds whose topology (e.g. shape, dimension, equilibria, etc.) may be coarsely captured by homological differences. Given these assumptions, and partially inspired by prior successes using topological features to differentiate between different motion types \cite{PHMotion} and dynamical systems \cite{PIs}, we describe below a novel approach for extracting topological features from motion tracks.
\begin{figure*}[htbp]
\centering
\includegraphics[width=.85\textwidth]{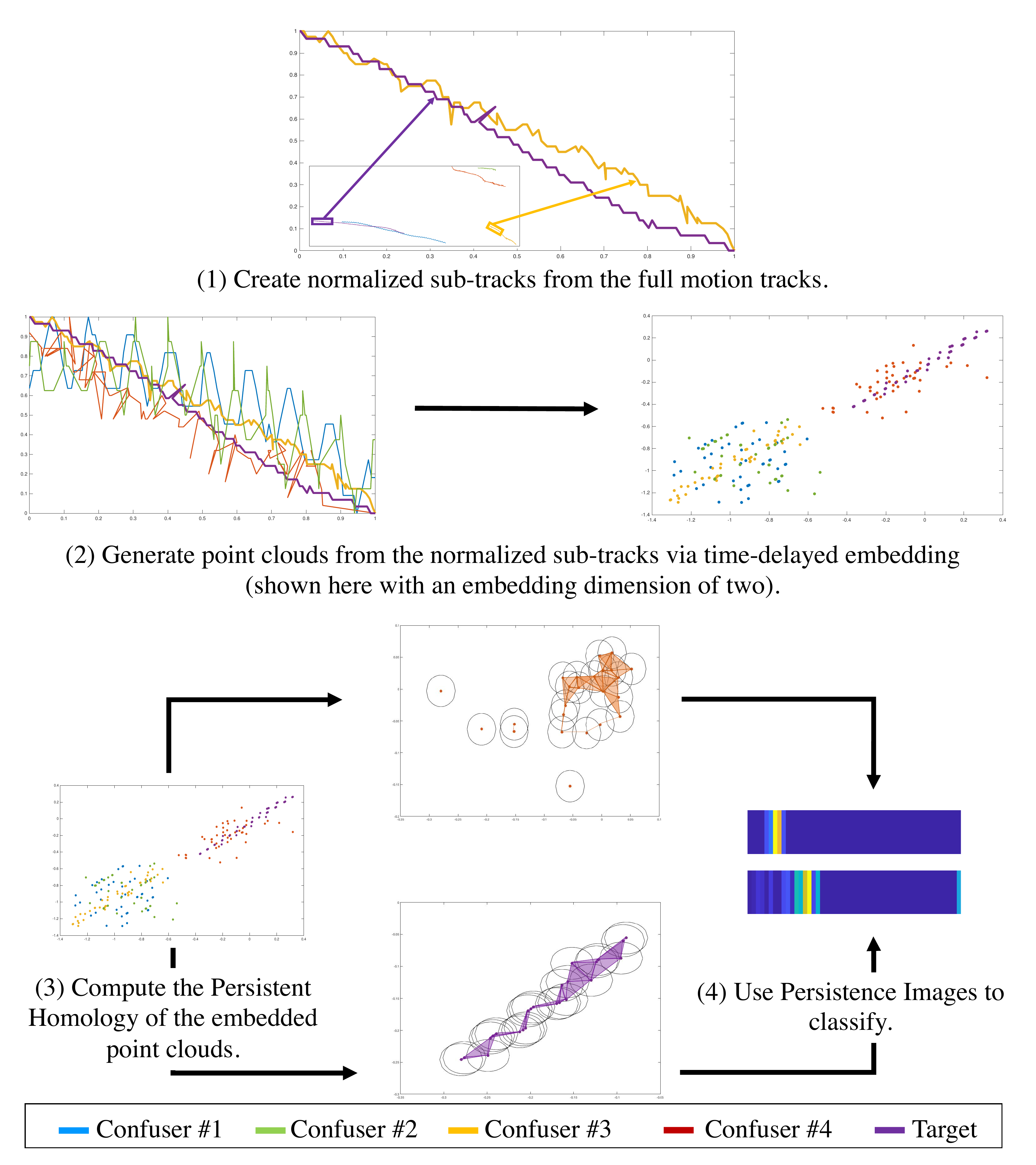}
\caption{Track classification loop: the persistent homology of the cloud of embedded track points is computed and then summarized using a persistence image. Here, we have shown the Vietoris-Rips complex at a single fixed scale arising from a confuser and a target subtrack.  }
\label{fig:graphicalAbstract}
\end{figure*}
\subsection{Experimental Design}
We assume that we begin with normalized sub-tracks of equal length. For each sub-track we perform a time-delayed embedding. For visualization sake we have embedded the data into the plane (i.e., we set the embedding dimension as $D=2$) using a time-delay of $\tau=1$. The Persistent Homology of each 2-D point cloud is computed using the Python Ripser package \cite{ripser}. For this first round of exploratory work we have chosen to look at the $0$-dimensional homology only. Once we have computed the homology we generate the corresponding persistence images (see Fig.~\ref{fig:graphicalAbstract}). For $0$-dimensional homology the persistence image is generally a vector and not an image. The reasoning for this is outside the scope of this paper but is tied to the fact that all $0$-dimensional homological features usually appear at the same scale (the null scale) and what differs is the scales at which they disappear. These labeled persistence vectors, $\vec{p}_k^{j}$, are the inputs to our classification algorithm.

Each of the $J$ objects in the scene yields a set, $P_j$, of $K_{j}$ persistence vectors which ultimately yields a superset $\mathcal{P}=\{P_{j}\}_{j=1}^{J}$ of persistence vectors. For this initial work we have implemented a $k$-nearest neighbors classifier using Python's SciKit-Learn \cite{scikit-learn}. We split each subset comprising the superset equally to yield training and test sets on which to test classification performance. We then explore the effect of track length on classification performance by testing subtracks consisting of 50, 75, and 100 measurements. As a comparison, we also apply $k-$nearest neighbors to the normalized sub-track statistics of the same length. Figure~\ref{fig:graphicalAbstract} shows an outline of the full track classification loop.

\section{Results}
\label{sec:experiment}
The results of the experiments described in Section \ref{sec:data} are presented in Table \ref{tbl:results}. ``Track Length" refers to the number of measurements used to generate a sub-track. ``Feature Method" indicates whether the $k-$nearest neighbors algorithm was applied to the 1-D statistic, $\vec{s}_k^{j}$, computed from the normalized sub-track or the $H_0$ persistence vector, $\vec{p}_k^{j}$, arising from embedding the statistic. Within the ``Confusion Matrix" column we indicate the true class label by sub-column heading and the class predicted by the $k-$nearest neighbors algorithm as sub-rows. In the following, as well as in Section \ref{sec:conclusion}, we associate the target and confuser classes with the notion of ``positive'' and ``negative'' classes, respectively.

For each of the three track lengths we see that the persistence vector of the embedded statistic is better able to differentiate between the two classes. At all three lengths the persistence vectors are able to correctly classify all of the confusers and it is not until the shortest length that we see the first misclassification of a target. Even at this shortest length the persistence vectors only have a $1\%$ false negative rate.

In the context of our motivating application a false negative is more acceptable than a false positive and both feature methods achieve at most a $2\%$ false negative rate. However, we see that the 1-D statistic produces high false positive rates that increase as the track length is decreased. Excessive false positives degrade our ability to extract targeted higher-resolution images for better classification at the rates required to aid with time- and distance-dependent response decisions.
\begin{table}[htbp]
\caption{Classification results using a $k$-Nearest Neighbors Classifier. Columns labels indicate true class row labels indicate the predicted class. The $k$-Nearest Neighbors classifier was trained using 50\% of the data and the reported results are for the remaining 50\% of the data.}\label{tbl:results}
\begin{center}
\begin{tabular}{|c|c|r|c|c|}
\hline
\textbf{Track}& \textbf{Feature} & \multicolumn{3}{|c|}{\textbf{Confusion Matrix}} \\
\textbf{Length} & \textbf{Method}& $\downarrow$\textbf{\textit{Predicted}}/\textbf{\textit{True }}$\rightarrow$& \textbf{\textit{Confuser}}& \textbf{\textit{Target}} \\
\hline
 & Statistic of& \textbf{\textit{Confuser}}&  0.91 & 0.09\\
100 & Normalized Subtrack & \textbf{\textit{Target}} & 0.00 & 1.00\\
\cline{2-5}
 & Persistence Vector of & \textbf{\textit{Confuser}}&  \textbf{1.00} & \textbf{0.00}\\
 & Embedded Statistic & \textbf{\textit{Target}} & \textbf{0.00} & \textbf{1.00}\\
\hline
 & Statistic of& \textbf{\textit{Confuser}}&  0.81 & 0.19\\
75 & Normalized Subtrack & \textbf{\textit{Target}} & 0.02 & 0.98\\
\cline{2-5}
 & Persistence Vector of& \textbf{\textit{Confuser}}&  \textbf{1.00} & \textbf{0.00}\\
 & Embedded Statistic & \textbf{\textit{Target}} & \textbf{0.00} & \textbf{1.00}\\
\hline
 & Statistic of& \textbf{\textit{Confuser}}&  0.76 & 0.24\\
50 & Normalized Subtrack & \textbf{\textit{Target}} & 0.02 & 0.98\\
\cline{2-5}
 & Persistence Vector of& \textbf{\textit{Confuser}}&  \textbf{1.00} & \textbf{0.00}\\
 & Embedded Statistic & \textbf{\textit{Target}} & \textbf{0.01} & \textbf{0.99}\\
\hline

\end{tabular}
\label{tab1}
\end{center}
\end{table}

\section{Conclusion}
\label{sec:conclusion}
We have implemented a novel framework for distinguishing between small targets and confusers based on analysis of motion tracks derived from wide-field-of-view images. Our approach leverages the rich theory of both time-delayed embeddings and topological data analysis. We tested our approach on a data set comprised of five motion tracks where one track is the positive (target) class and the other four belong to the negative (confuser) class. We find that the topological features are a strong discriminator between the classes and are robust to sub-track length.

In future work we will perform further benchmarking against alternative approaches. Additionally, we will utilize more sophisticated methods of selecting time-delay embedding parameters not implemented in this initial, proof-of-concept work. We also plan to expand the set of homological dimensions considered and apply deep learning image classifiers to the resulting persistence images.

\bibliographystyle{plain}
\bibliography{MotionDiscrimination.bib}

\end{document}